\documentclass[journal,twoside,web]{ieeecolor}
\usepackage{generic}
\usepackage{cite}
\usepackage{amsmath,amssymb,amsfonts}
\usepackage{algorithmic}
\usepackage{graphicx}
\usepackage{textcomp}
\usepackage{hyperref}
\hypersetup{hidelinks}
\usepackage{booktabs} % 三线表
\usepackage{multirow} % 表格多行合并
\usepackage{siunitx}  % 表格数值对齐
\usepackage{url}
\usepackage{stfloats} % 优化双栏底部浮动体

\def\BibTeX{{\rm B\kern-.05em{\sc i\kern-.025em b}\kern-.08em
		T\kern-.1667em\lower.7ex\hbox{E}\kern-.125emX}}

\begin{document}
	
	\title{Anatomy Aware Cascade Network: Bridging Epistemic Uncertainty and Geometric Manifold for 3D Tooth Segmentation}
	
	\author{Bing Yu, Liu Shi, Haitao Wang, Deran Qi, Xiang Cai, Wei Zhong, Qiegen Liu, \IEEEmembership{Senior Member, IEEE}
		\thanks{This work was supported by the National Natural Science Foundationof China(Grant:621220033 and 62201193), Nanchang University YouthTalent Training Innovation Fund Project(Grant:XX202506030012), Early-Stage Young Scientific and Technological Talent Training Foundation ofJiangxi Province (Grant:KJ202509220191). (Bing Yu and Liu Shi are the co-first authors.) (Cocorresponding author: Liu Shi.)}
		\thanks{Liu Shi, Qiegen Liu, Bing Yu, Deran Qi are with the School of Information Engineering, Nanchang University, Nanchang 330031, China. (e-mail: {shiliu,liugiegen}@ncu.edu.cn; yubing@email.ncu.edu.cn; 6108123121@email.ncu.edu.cn).}
		\thanks{Haitao Wang is with the First Affiliated Hospital, Jiangxi Medical College, Nanchang University, Nanchang 330031, China. (e-mail: wanghaitao20000103@163.com).}
		\thanks{Xiang Cai is with The Second Clinical Medical College, Nanchang University, Nanchang 330031, China (e-mail: shuixiang66@qq.com).}
		\thanks{Wei Zhong is with The First Clinical Medical College, Nanchang University, Nanchang 330031, China (e-mail: zw15279605261@outlook.com).}
	}
	
	\maketitle
	
	\begin{abstract}
		Accurate three-dimensional (3D) tooth segmentation from Cone-Beam Computed Tomography (CBCT) is a prerequisite for digital dental workflows. However, achieving high-fidelity segmentation remains challenging due to adhesion artifacts in naturally occluded scans, which are caused by low contrast and indistinct inter-arch boundaries. To address these limitations, we propose the Anatomy Aware Cascade Network (AACNet), a coarse-to-fine framework designed to resolve boundary ambiguity while maintaining global structural consistency. Specifically, we introduce two mechanisms: the Ambiguity Gated Boundary Refiner (AGBR) and the Signed Distance Map guided Anatomical Attention (SDMAA). The AGBR employs an entropy based gating mechanism to perform targeted feature rectification in high uncertainty transition zones. Meanwhile, the SDMAA integrates implicit geometric constraints via signed distance map to enforce topological consistency, preventing the loss of spatial details associated with standard pooling. Experimental results on a dataset of 125  CBCT volumes demonstrate that AACNet achieves a Dice Similarity Coefficient of  90.17 \% and a 95\% Hausdorff Distance of 3.63 mm, significantly outperforming state-of-the-art methods. Furthermore, the model exhibits strong generalization on an external dataset with an HD95 of 2.19 mm, validating its reliability for downstream clinical applications such as surgical planning. Code for AACNet is available at \url{https://github.com/shiliu0114/AACNet}. 
		
	\end{abstract}
	
	
	\section{Introduction}
	\label{sec:introduction}
	\IEEEPARstart{C}{one-beam} Computed Tomography (CBCT) has become an indispensable three-dimensional (3D) imaging modality in modern digital dentistry, favored for its high spatial resolution and relatively low radiation dose. Precise 3D tooth models form the foundation of advanced clinical applications, ranging from orthodontic treatment planning \cite{mozzo1998new,merrett2009cone,hechler2008cone}, and implant surgery to endodontic navigation\cite{scarfe2006clinical,angelopoulos2008comparison,benavides2012use}. Consequently, automatic tooth instance segmentation is a crucial yet challenging task in constructing clinically actionable digital models.
	
\begin{figure}[!t]
		\centering
		\includegraphics[width=1.0\columnwidth]{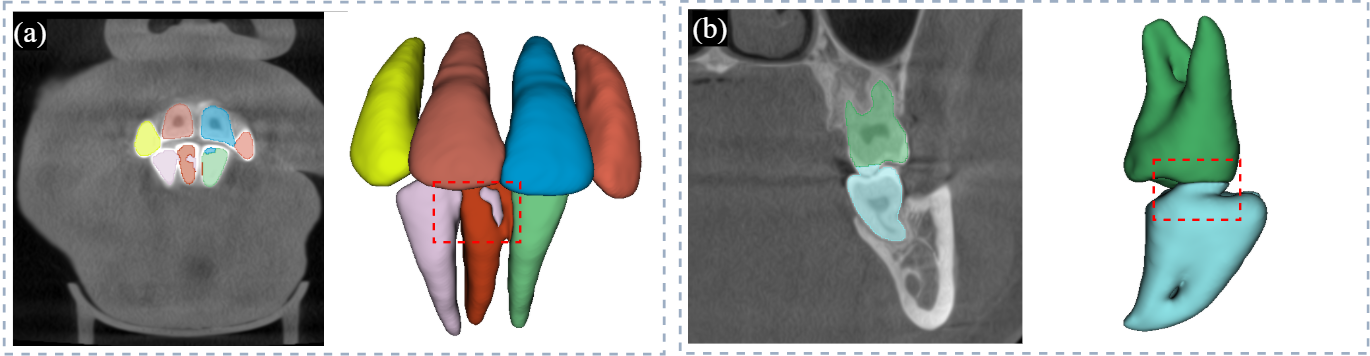}
		\caption{Illustration of typical adhesion artifacts. (a) Inter proximal adhesion between adjacent teeth. (b) Inter arch adhesion at the occlusal interface. In both scenarios, low grayscale contrast at contact zones leads to severe boundary ambiguity and erroneous fusion in conventional segmentation.}
		\label{fig:fig1_1}
	\end{figure}

	Despite recent advancements, tooth segmentation from CBCT data faces persistent intrinsic challenges. First, as illustrated in Fig. \ref{fig:fig1_1}, in naturally occluded scans, the contact between maxillary and mandibular teeth results in minimal grayscale variation at the occlusal surface \cite{hosntalab2008segmentation,ji2014level}, severely obscuring the inter-arch boundaries. Second, the intensity distribution of tooth roots often mimics that of the surrounding alveolar bone, blurring structural interfaces. This is a common artifact in medical imaging that complicates boundary detection. Third, the high morphological similarity among adjacent teeth can lead to instance misidentification \cite{miki2017classification,tuzoff2019tooth}. These factors collectively cause conventional segmentation methods, which rely heavily on voxel intensity and gradient information to fail \cite{kendall2017uncertainties,wang2019aleatoric}.
	
	% Figure 1
	
	Specifically, low-level techniques such as intensity based thresholding and region growing are highly sensitive to low-contrast conditions. These methods often fail to establish robust stopping criteria at the occlusal plane, resulting in over segmentation and the erroneous merging of dental arches \cite{nackaerts2015segmentation,akhoondali2009rapid,keyhaninejad2006automated}. Similarly, complex approaches like template based fitting lack robustness when facing significant inter patient variations in tooth shape \cite{barone2016ct}. While level-set-based methods have demonstrated superior performance , they are typically semi-automatic, requiring manual initialization of contours or seed points \cite{xia2017individual},\cite{gao2010individual}. Consequently, these traditional methods often fail to preserve individual tooth topology and require extensive manual post-processing, rendering them unsuitable for fully automated clinical workflows.
	
	The advent of deep learning has significantly advanced medical image segmentation, with Convolutional Neural Networks (CNNs) and transformer architectures demonstrating significant potential. Numerous studies have applied these techniques to automatic tooth segmentation. Xu \textit{et al}. \cite{xu20183d} proposed a two-layer CNN for CT scans to label gingiva and inter-dental spaces. Lee \textit{et al}. \cite{lee2020automated} and Rao \textit{et al}. \cite{rao2020symmetric} utilized Fully Convolutional Networks \cite{long2015fully} for whole-tooth segmentation, while Cui \textit{et al}. \cite{cui2019toothnet} developed a Mask R-CNN-based framework \cite{he2017mask} for individual tooth identification. More recently, Zhou \textit{et al}. \cite{zhou2023nnformer} introduced nnFormer, a 3D Transformer based model leveraging local and global self-attention for volumetric representation. Additionally, Zhong \textit{et al}. \cite{zhong2025pmfsnet} proposed the lightweight PMFSNet for CBCT segmentation, and Lyu \textit{et al}. \cite{lyu2024crml} addressed segmentation via cross-modal inference and multi-task learning.
	
	Despite these successes, limitations remain. CNNs are constrained by local receptive fields, while transformers face high computational complexity. Consequently, current methods still struggle with issues such as edge blurring due to intensity similarity, distinguishing adjacent teeth, and managing the interaction between teeth and surrounding tissues.
	
	To overcome these limitations, we propose the Anatomy Aware Cascade Network (AACNet). This framework deeply couples uncertain boundary modeling with implicit geometric regularization via a two-stage cascade strategy. First, to address texture ambiguity in adhesion zones, we introduce the Ambiguity Gated Boundary Refiner (AGBR). This module employs an ambiguity field, derived from a Shannon entropy approximation, to explicitly quantify voxel level epistemic uncertainty. It acts as a dynamic gating signal, forcing the network to perform targeted feature rectification in high-entropy transition zones, thereby enhancing boundary discrimination. Second, to resolve geometric discontinuities and topological errors, we employ the Signed Distance Map guided Anatomical Attention (SDMAA) module. This module replaces traditional global average pooling with a surface integral on the anatomical manifold, using the Lipschitz-continuous signed distance map (SDM) as an anatomical prior. By directly embedding implicit geometric gradient information into the latent feature space, SDMAA significantly enhances topological consistency and ensures robustness in weak texture regions.
	
	In summary, the main contributions of this study are as follows:
	\begin{itemize}
		\item We propose a coarse-to-fine segmentation framework that decomposes the complex 39-class tooth segmentation task, explicitly addressing the boundary ambiguity between maxillary and mandibular arches.
		
		\item Bridging Uncertainty and Geometry Gaps:
		The AGBR leverages a Gini-derived ambiguity field to perform targeted feature rectification in high-uncertainty regions, effectively improving boundary delineation at inter-dental contacts and tooth–bone interfaces.
		Complementarily, the SDMAA integrates Lipschitz-continuous geometric priors into the feature aggregation process, promoting topological consistency and reducing structural discontinuities and adhesion artifacts in fine dental structures.
		Synergistically, these modules jointly enhance boundary accuracy and anatomical plausibility, leading to more reliable dental segmentation for downstream clinical applications.
		
		\item We bridge the gap between algorithmic development and clinical application by validating the model on real-world preoperative assessment tasks, specifically sinus and nerve proximity analysis, demonstrating its reliability for surgical planning.
	\end{itemize}
	
	\section{Related Work}
	\subsection{Deep Learning for Medical Image Segmentation}
	The field of medical image segmentation has evolved rapidly from CNNs to Transformer-based architectures. The canonical U-Net \cite{ronneberger2015u} and its volumetric variants like 3D U-Net \cite{cciccek20163d} and nnU-Net \cite{isensee2021nnu} established strong baselines by learning voxel-level spatial contexts. To overcome the limited receptive fields of CNNs, Transformer-based models such as UNETR \cite{hatamizadeh2022unetr}, nnFormer \cite{zhou2023nnformer}, and hybrid architectures like 3DUX-Net \cite{lee20223d} were introduced to capture global dependencies.
	Driven by these advancements, numerous studies have applied these techniques specifically to dental segmentation. Cui \textit{et al}. \cite{cui2019toothnet} utilized a 3D Region Proposal Network for instance segmentation, while Wu \textit{et al}. \cite{wu2020center} adopted a hierarchical approach for tooth center identification. More recently, Cui \textit{et al}. \cite{cui2022fully} achieved high accuracy on large-scale datasets, and Chen \textit{et al}. \cite{chen2023cta} proposed a hybrid CNN-Transformer architecture for dental CBCT. However, these methods primarily focus on architectural improvements rather than explicitly addressing the unique geometric challenges of dental adhesion.
	
	\subsection{Uncertain Boundary Modeling}
	Accurate boundary delineation is critical for clinical applications. Recent studies have introduced explicit boundary branches or attention modules. For instance, BA-Net \cite{zhou2022banet} uses a dedicated edge channel, while EBA-Net \cite{jia2022ebanet} employs a dual attention mechanism. Huang \textit{et al}. \cite{huang2025lesion} utilized edge priors to guide attention, and Jiang \textit{et al}. \cite{jiang2026e2miseg} proposed E2MISeg with scale-sensitive loss for blurred regions. However, most existing methods apply a uniform refinement strategy, lacking adaptive mechanisms to selectively focus on uncertain or ambiguous regions, which is crucial for distinguishing fused tooth roots.
	
	\subsection{Implicit Geometric Regularization}
	In medical images with low contrast, relying solely on texture often fails to ensure anatomical plausibility. Researchers have addressed this by introducing shape priors. The Anatomically Constrained Neural Network \cite{oktay2017anatomically} uses an autoencoder to learn a latent shape manifold. Similarly, You \textit{et al}. \cite{you2024learning} introduced a Shape Prior Module, and Shi \textit{et al}. \cite{shi2023nextou} proposed NexToU using Graph Neural Networks. Zhu \textit{et al}. \cite{zhu2025improving} employed the signed distance field as an auxiliary task. However, most prior works impose shape regularization externally via loss functions rather than integrating shape priors directly into the feature learning process.
	
	While progress has been made in boundary refinement and shape consistency, these approaches have developed largely in parallel. Boundary-centric methods often lack global anatomical constraints, leading to topological errors. Conversely, shape prior methods apply global regularization but often fail to address local high-frequency blurring. Consequently, a comprehensive framework addressing boundary ambiguity, inter-arch adhesion, and anatomical consistency simultaneously is lacking. The AACNet proposed in this paper aims to bridge this gap by integrating ambiguity-guided boundary refinement with SDM-based anatomical priors.
	
	% Figure 2
	\begin{figure}[!t]
		\centering
		\includegraphics[width=\linewidth]{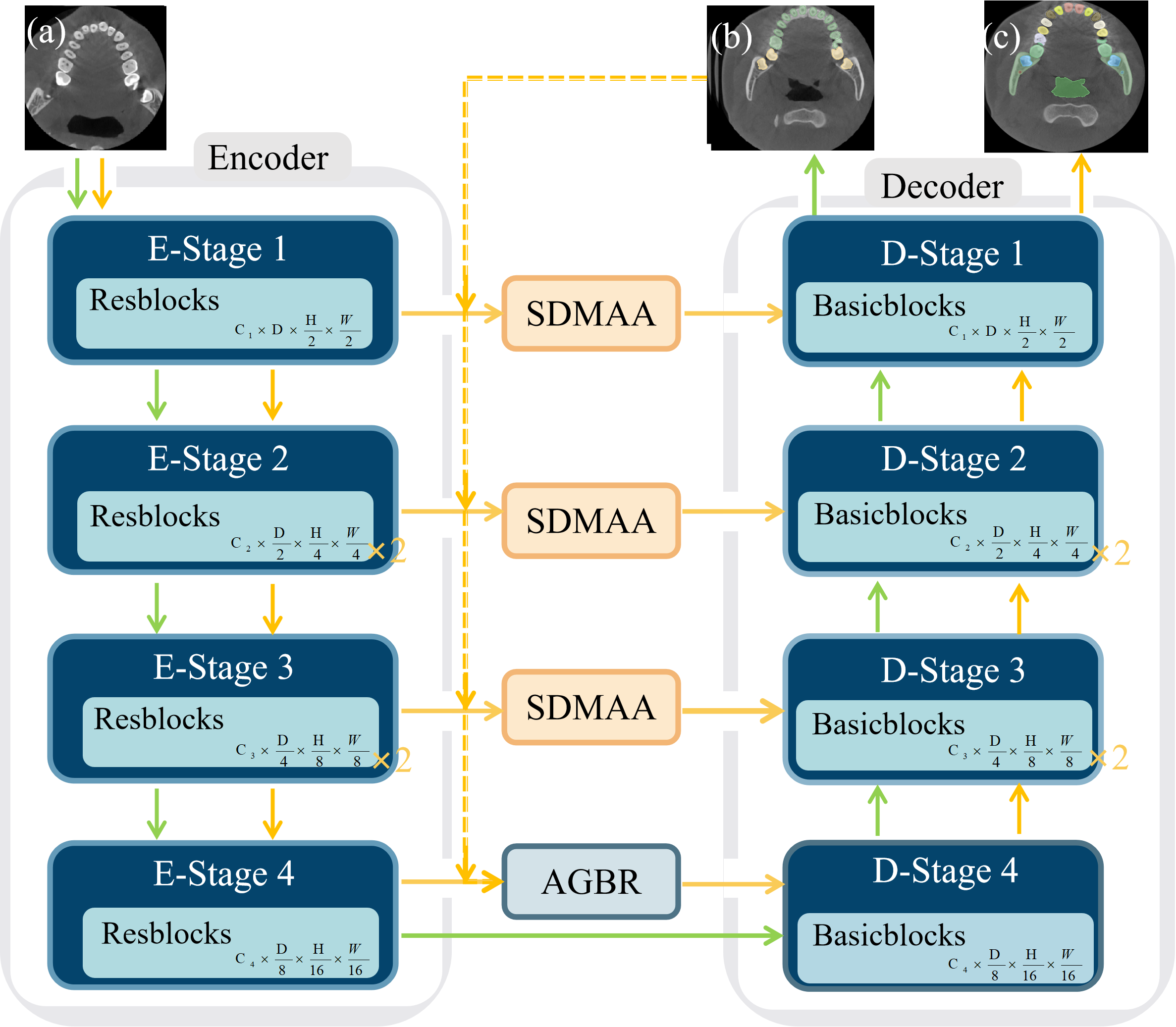}
		\caption{Overview of the proposed AACNet framework. The method adopts a coarse-to-fine decomposition strategy where the green line illustrates Stage I generating a global ambiguity field from the input volume, shown as (a) the original CBCT, resulting in the probabilistic priors displayed as (b) the coarse localization maps ($P_{upper}, P_{lower}$). Subsequently, the yellow line depicts Stage II fusing these priors with the original image. By utilizing two key mechanisms, AGBR and SDMAA, the network hierarchically addresses boundary ambiguity and enforces topological consistency, yielding (c) the final segmentation result.}
		\label{fig:aacn}
	\end{figure}

	\begin{figure}[!t]
		\centering
		\includegraphics[width=\linewidth]{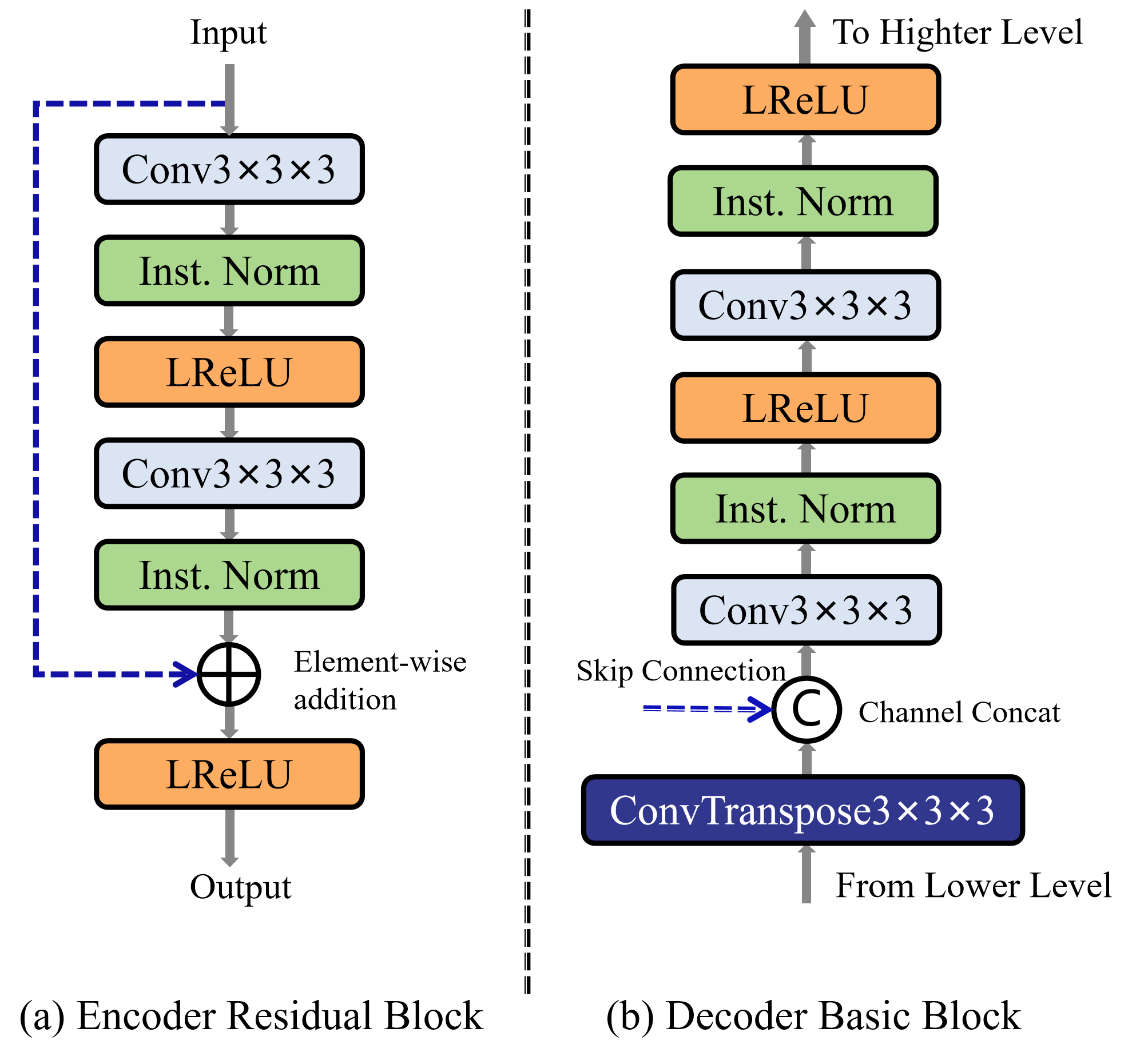}
		\caption{Detailed architecture of the fundamental building blocks used in the Residual U-Net backbone. (a) The Encoder Residual Block, featuring stacked convolutional layers with Instance Normalization and LeakyReLU to preserve feature identity during deep extraction. (b) The Decoder Basic Block, designed for spatial recovery, utilizing transposed convolutions for upsampling followed by feature concatenation and refinement convolutions.}
		\label{fig:network_blocks}
	\end{figure}

	\section{Methodology}
	\label{sec:methodology}
	
	We propose the AACNet to address adhesion artifacts and boundary ambiguity. As illustrated in Fig. \ref{fig:aacn}, the framework adopts a coarse-to-fine decomposition. Stage I generates global probabilistic priors, which guide Stage II. Central to this architecture, the AGBR and SDMAA modules are synergistically integrated into a Residual U-Net to hierarchically resolve boundary uncertainty and strictly enforce anatomical topology.
	
	\subsection{Coarse-to-Fine Decomposition Framework}
	Conventional end-to-end networks typically attempt to segment all 39 classes simultaneously, a strategy that often fails at the highly ambiguous occlusal plane. In contrast, our AACNet decouples this problem into two sub-tasks: Stage I for coarse localization and Stage II for fine-grained refinement.
	
	To balance efficiency and representation, we adopt a unified Residual U-Net backbone. As detailed in Fig. \ref{fig:network_blocks}, the encoder utilizes stacked Residual blocks to preserve feature identity during deep extraction. Conversely, the decoder employs basic blocks consisting of transposed convolutions for upsampling and concatenation operations to recover spatial resolution and fuse semantic features.
	
	\subsubsection{Stage I: Coarse Localization and Ambiguity Field Generation}
	Let $I \in \mathbb{R}^{D \times H \times W}$ denote the input CBCT volume. The primary objective of Stage I is to generate a simplified binary localization of the maxillary and mandibular arches, rather than a fine grained segmentation. Given that this stage prioritizes rapid Region of Interest extraction over fine detail, we employ a lightweight configuration of the backbone described above. This variant utilizes fewer residual blocks and reduced channel dimensions to minimize memory consumption and inference latency, while maintaining sufficient capacity to generate the global ambiguity field. This process is formulated as:
	\begin{equation}
		(P_{upper}, P_{lower}) = \mathcal{F}_{StageI}(I),
	\end{equation}
	where $\mathcal{F}_{StageI}$ represents the Stage I network, and $P_{upper}, P_{lower} \in [0, 1]^{D \times H \times W}$ denote the probability maps for the upper and lower dental arches, respectively. These maps collectively constitute an ambiguity field, which serves as a spatial prior for the subsequent stage.
	
	Crucially, Stage I does not demand pixel-perfect accuracy. rather, it provides an approximate region of interest and a probability landscape. This cascading design ensures robustness: even with coarse boundaries from Stage I, the generated ambiguity field explicitly highlights uncertain areas. This guides Stage II to focus specifically on rectifying these errors via the AGBR module. Notably, despite being a coarse stage, our lightweight localization network achieves a Dice Similarity Coefficient (DSC) of over 95\% on the validation set. This high-fidelity localization ensures the ambiguity field serves as a reliable spatial prior for the subsequent fine segmentation.
	
	\subsubsection{Stage II: Prior-Guided Anatomical Fine-Segmentation}
	The second network, $\mathcal{F}_{StageII}$, is responsible for the final 39-class fine grained segmentation. To capture intricate dental features and resolve boundary ambiguities, we instantiate the backbone with a deep configuration comprising 6 resolution levels. Specifically, the encoder is significantly deepened, with the number of residual blocks in each stage set to $[1, 3, 4, 6, 6, 6]$, respectively. Feature channels are progressively increased as $[32, 64, 128, 256, 320, 320]$ to enrich semantic representation. The decoder employs transposed convolutions for upsampling and integrates features from the encoder via skip connections. Furthermore, deep supervision is applied at five resolution scales to ensure robust gradient propagation across this deep architecture.
	
	The input $I_{in}$ is a multi-channel tensor formed by concatenating the original image $I$ with the probability priors calibrated by a convolutional adapter $\mathcal{A}$:
	\begin{equation}
		I_{in} = \text{Concat}(I, \mathcal{A}(\text{Concat}(P_{upper}, P_{lower}))),
	\end{equation}
	where the adapter $\mathcal{A}$ is a $1 \times 1 \times 1$ convolution designed to align the feature space of the probability priors with that of the CBCT image. The backbone of $\mathcal{F}_{StageII}$ is a residual U-Net, within which we deeply integrate the AGBR and SDMAA modules to produce the final segmentation prediction $\hat{Y}$.
	
	\subsection{Modeling Uncertainty via Ambiguity Gating}
	\label{subsec:agbr}
	
	\begin{figure}[!t]
		\centering
		\includegraphics[width=\linewidth]{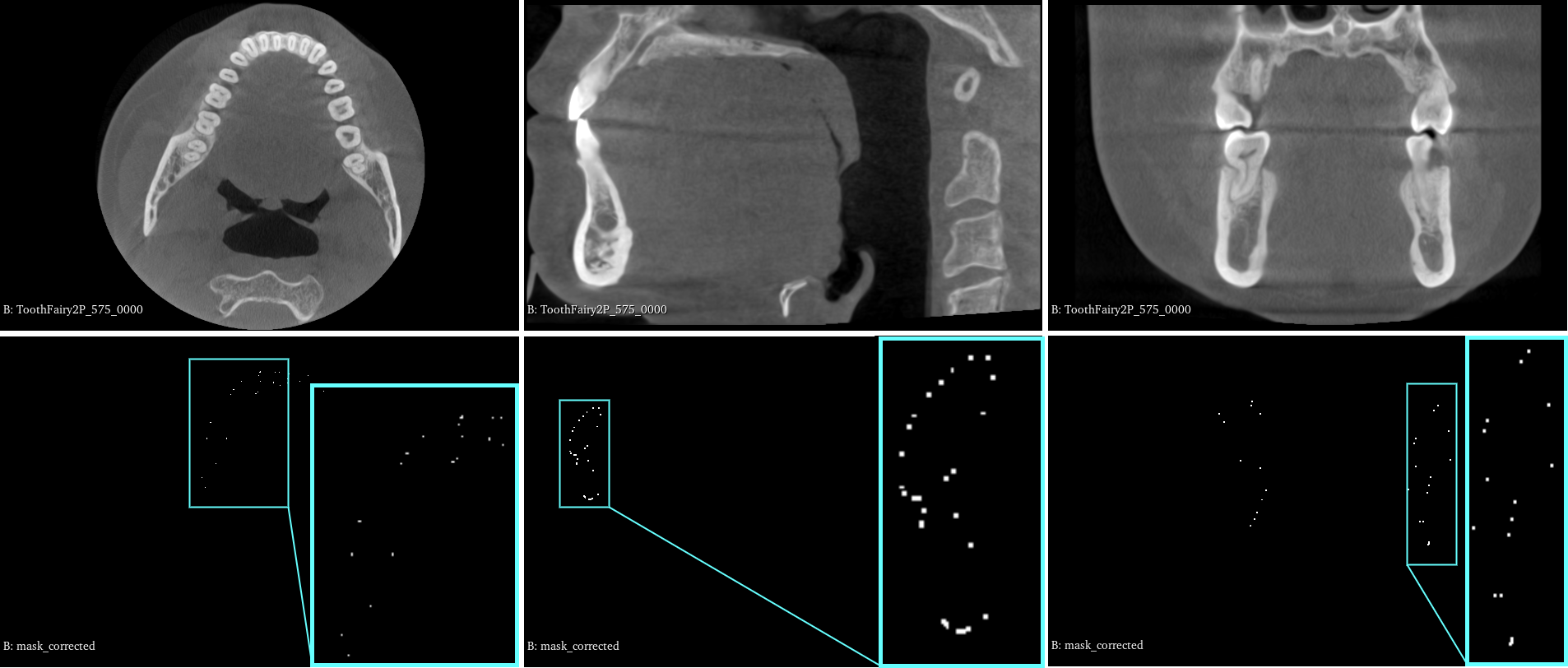}
		\caption{Visualization of the ambiguity field \(A(v)\) derived from Stage I predictions. Bright regions indicate voxels of maximal epistemic uncertainty, which correspond precisely to the blurred occlusal interfaces and inter proximal contact zones. By applying a high threshold ($\tau=0.95$), we generate a sparse binary gating mask from this field, directing the AGBR module to focus refinement efforts exclusively on these hardest-to-segment boundaries.}
		\label{fig:agbr_mask}
	\end{figure}
	
	\begin{figure}[!t]
		\centering
		\includegraphics[width=\linewidth]{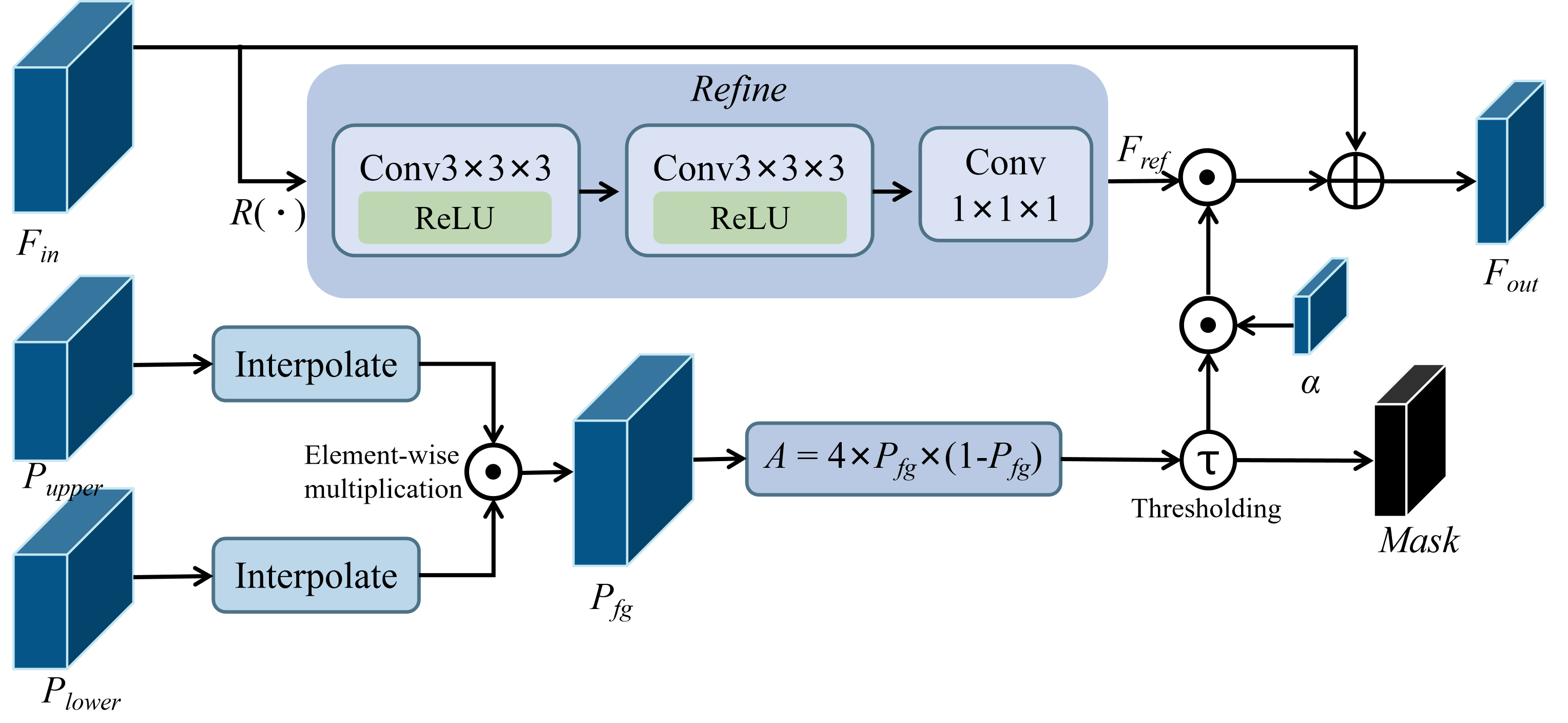}
		\caption{Detailed of the AGBR module. The module quantifies voxel-level epistemic uncertainty from Stage I predictions to generate a binary gating mask via thresholding. This mask acts as a dynamic switch, activating the residual refinement branch $\mathcal{R}(\cdot)$ exclusively in high-entropy transition zones. This mechanism performs targeted feature rectification on ambiguous boundaries while preserving the integrity of high-confidence regions.}
		\label{fig:agbr1-8}
	\end{figure}

	Standard segmentation networks typically treat all voxels with equal importance, neglecting the intrinsic variation in predictive difficulty across the volumetric domain. In dental CBCT, naturally occluded regions exhibit severe epistemic uncertainty due to low contrast and indistinct inter-arch boundaries, whereas internal tooth structures are relatively well  defined. As illustrated in Fig. \ref{fig:agbr1-8}, we introduce the AGBR to resolve this disparity, serving as an entropy-aware gate. Unlike conventional operations that perform uniform processing, the AGBR explicitly leverages the probability dispersion from the coarse stage to quantify voxel level ambiguity. This allows the network to dynamically perform targeted feature rectification specifically within the high uncertainty transition zones, bridging the gap between confident anatomical structures and ambiguous contact interfaces.
	
	The process begins with ambiguity quantification. Leveraging the probability priors generated in Stage I, we define the foreground probability $P_{fg}$ and the derived ambiguity field $A$. For any given voxel $v$:
	\begin{equation}
		P_{fg}(v) = \frac{P_{upper}(v) + P_{lower}(v)}{2},
	\end{equation}
	\begin{equation}
		A(v) = 4 \cdot P_{fg}(v) \cdot (1 - P_{fg}(v)).
		\label{eq:ambiguity}
	\end{equation}
	Mathematically, Equation 4 constitutes a normalized uncertainty metric derived from the Gini impurity index. While Shannon entropy is a standard measure for uncertainty, it necessitates logarithmic computations that incur higher computational overhead and potential numerical instability near zero probabilities. In contrast, the quadratic form provides smooth, well-behaved gradients around the decision boundary where $P$ approximates $0.5$, facilitating stable backpropagation. The scaling factor of 4 normalizes the metric to the range $[0, 1]$, ensuring that $A(v)=1$ represents maximal epistemic uncertainty, occurring precisely when $P_{fg}$ equals $0.5$.
	
	Based on this field, we generate a binary gating mask $M(v)$ to explicitly isolate these challenging voxels:
	\begin{equation}
		M(v) = \mathbb{I}(A(v) > \tau),
	\end{equation}
	where $\mathbb{I}(\cdot)$ denotes the indicator function, and $\tau$ is a predefined ambiguity threshold. Statistical analysis of the ambiguity distribution reveals a significant skew towards zero, indicating high confidence, with a long tail representing uncertain regions. In this study, we set $\tau = 0.95$. This high threshold targets the extreme tail of the distribution, corresponding to the prediction probability band of approximately $[0.39, 0.61]$. As a result, the AGBR is strictly activated only for the most equivocal inter-arch boundaries, while the vast majority of confident regions remain unaffected. This strategy prevents the corruption of well-learned features in high-confidence regions while concentrating refinement capacity on the ambiguous topology.
	
	The core of the AGBR lies in its uncertainty-driven gated residual refinement. Let $F_{in} \in \mathbb{R}^{C \times D' \times H' \times W'}$ denote the input feature map. The module processes $F_{in}$ via two parallel paths: an identity path and a gated refinement path. The latter is governed by a lightweight bottleneck sub-network $\mathcal{R}(\cdot)$:
	\begin{equation}
		\begin{split}
			F_{ref} = \mathcal{R}(F_{in}) = \text{Conv}_{1\times1\times1} \big( \mathrm{ReLU} ( & \text{Conv}_{3\times3\times3} ( \mathrm{ReLU} ( \\
			& \text{Conv}_{3\times3\times3}(F_{in}) ) ) ) \big),
		\end{split}
	\end{equation}
	where the first $\text{Conv}_{3 \times 3 \times 3}$ layer projects the high-dimensional input features into a low-dimensional latent space to reduce computational overhead. The subsequent $\text{Conv}_{3 \times 3 \times 3}$ layer performs deep context extraction to rectify ambiguous boundary features. Finally, the $\text{Conv}_{1 \times 1 \times 1}$ layer restores the channel dimensions to match the input $F_{in}$, facilitating the element-wise residual addition.
	Finally, the output feature $F_{out}$ is obtained via a learnable residual fusion:
	\begin{equation}
		F_{out}(v) = F_{in}(v) + \alpha \cdot (M(v) \odot F_{ref}(v)),
	\end{equation}
	where $\odot$ denotes the element wise product and $\alpha$ is a learnable scaling factor initialized to 0. This formulation ensures that the backbone features are preserved in confident regions, while ambiguous regions receive a targeted, additive correction. We deploy the AGBR at the network's bottleneck to resolve critical ambiguities at the highest semantic level.
	
	\subsection{Enforcing Topology via Geometric Priors}
	\label{subsec:sdmaa}
	
	\begin{figure}[!t]
		\centering
		\includegraphics[width=\linewidth]{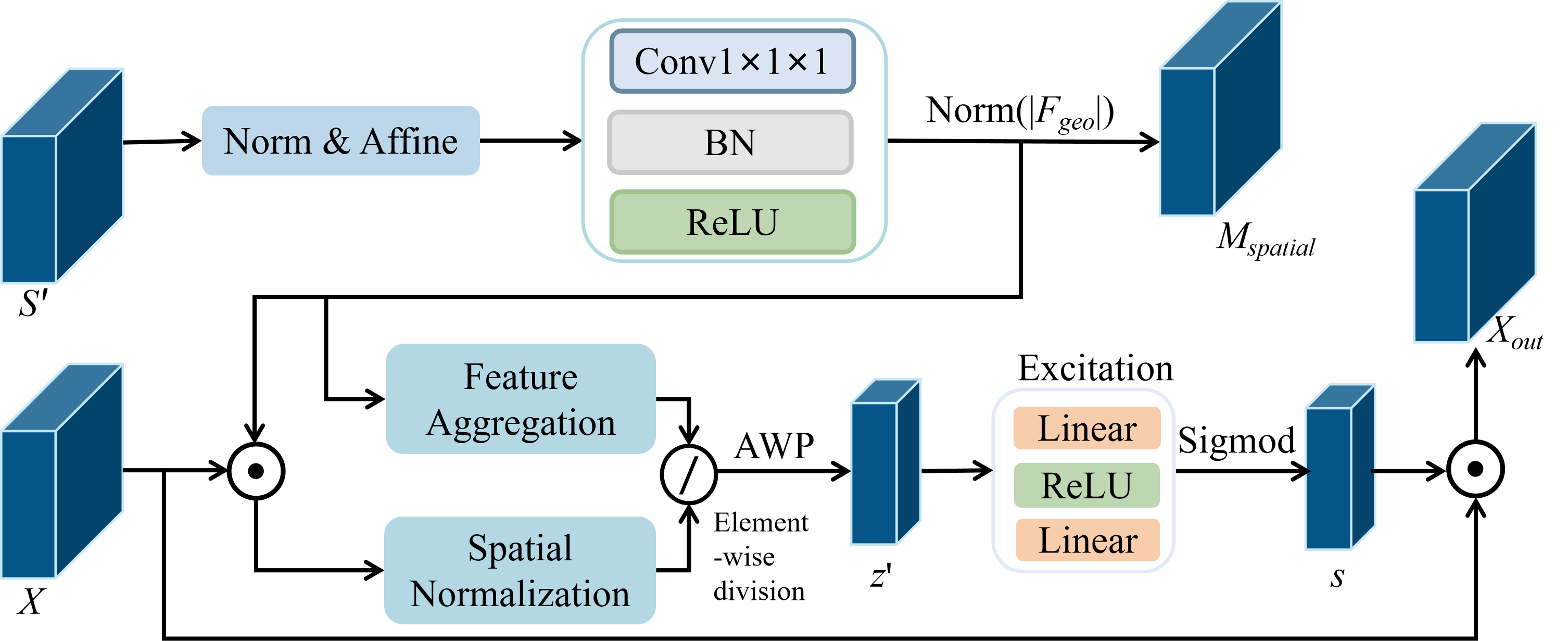}
		\caption{Detailed of the SDMAA module. The module embeds implicit geometric constraints into the latent feature space by: (1) Projecting the SDM into a spatial attention map $M_{spatial}$ via a learnable geometric adapter, and (2) Aggregating features using AWP. This process enforces the learned representations to align with the anatomical manifold, ensuring topological consistency in the final segmentation.}
		\label{fig:sdmaa1}
	\end{figure}

	\begin{figure}[!t]
		\centering
		\includegraphics[width=\linewidth]{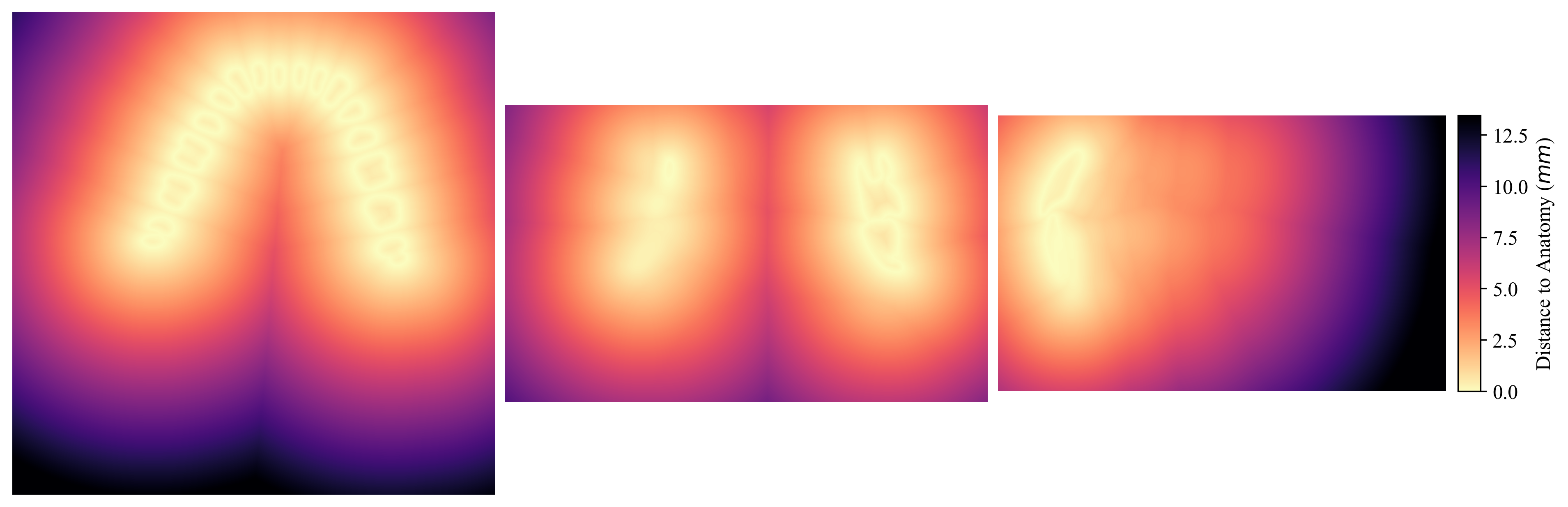}
		\caption{Visualization of the SDM used in the SDMAA module. The heatmap indicates the Euclidean distance to the nearest dental surface. Brighter regions highlight the anatomical manifold, demonstrating how the geometric prior guides the network to focus on the dental topology.}
		\label{fig:sdm_viz}
	\end{figure}

	Ensuring anatomical plausibility, particularly the topological continuity of tooth roots, is paramount for clinical utility yet poses a significant challenge for standard feature aggregation methods. While conventional attention mechanisms effectively model inter-channel dependencies, they typically rely on Global Average Pooling (GAP). This reliance inherently leads to the loss of spatial information, sacrificing geometric cues that are essential for boundary delineation. To address these limitations, we propose the SDMAA module shown in Fig. \ref{fig:sdmaa1}. This mechanism moves beyond simple feature re-weighting by embedding implicit geometric manifold constraints directly into the feature space. By utilizing the Lipschitz-continuous SDM as a structural prior, SDMAA enforces the refined representations to strictly adhere to the underlying anatomical topology, thereby preventing boundary irregularities common in weak texture regions.
	
	\subsubsection*{Mathematical Formulation of Geometric Prior}
	Formally, let $\Omega \subset \mathbb{R}^3$ denote the volumetric image domain. Consider a target dental structure represented by a subset of voxels $\mathcal{S} \subset \Omega$, where the boundary of this structure, representing the tooth surface, is denoted as $\partial \mathcal{S}$. The SDM, $\phi(\mathbf{v})$, is defined as the minimum Euclidean distance from any voxel $\mathbf{v} \in \Omega$ to the closest point on the boundary $\partial \mathcal{S}$, with the sign indicating inclusion:
	\begin{equation}
		\phi(\mathbf{v}) = \begin{cases} 
			-\min_{\mathbf{y} \in \partial \mathcal{S}} \|\mathbf{v} - \mathbf{y}\|_2 & \text{if } \mathbf{v} \in \mathcal{S}, \\
			0 & \text{if } \mathbf{v} \in \partial \mathcal{S}, \\
			\min_{\mathbf{y} \in \partial \mathcal{S}} \|\mathbf{v} - \mathbf{y}\|_2 & \text{if } \mathbf{v} \notin \mathcal{S},
		\end{cases}
		\label{eq:sdm_calc}
	\end{equation}
	where $\|\cdot\|_2$ denotes the $L_2$ norm. Unlike binary segmentation masks $\mathcal{M} \in \{0, 1\}$, which are discrete and exhibit zero gradients almost everywhere except at the boundary, the SDM offers critical theoretical advantages for enforcing topological consistency. 
	
	Primarily, $\phi(\mathbf{v})$ satisfies the Eikonal equation constraint $|\nabla \phi(\mathbf{v})| = 1$ almost everywhere, ensuring that the geometric representation is Lipschitz-continuous. This property provides a smooth and differentiable manifold that mitigates the gradient vanishing problems associated with binary cross-entropy near sharp boundaries. Furthermore, the SDM establishes a global geometric gradient flow. In contrast to binary masks that provide only local supervision, $\phi(\mathbf{v})$ and its gradient $\nabla \phi(\mathbf{v})$ encode the direction and magnitude required to reach the anatomical boundary from any spatial location, as visualized in Fig. \ref{fig:sdm_viz}. By embedding this field into the SDMAA module, we provide the network with explicit spatial guidance, compelling features to align with the underlying anatomical skeleton rather than relying solely on local texture cues.
	
	\subsubsection*{Dynamic Shape Prior Generation}
	A critical challenge in cascade networks is the potential distribution shift and geometric inaccuracies inherent in the coarse predictions from the first stage. Relying on a rigid mathematical function (e.g., exponential decay) to generate attention weights assumes a perfect distance field, which may lead to error propagation if the coarse boundary is imprecise. 
	
	To address this, we use a learnable geometric adapter module. Instead of manually crafting the attention distribution, we employ a lightweight convolutional module to project the SDM into the feature space. This allows the network to adaptively calibrate the shape prior and learn the optimal attention distribution for boundary delineation.
	
	Formally, let $S' \in \mathbb{R}^{1 \times D \times H \times W}$ denote the interpolated shape prior derived from Stage I. First, to ensure numerical stability across different scans, we apply instance normalization followed by a learnable affine transformation:
	\begin{equation}
		\tilde{S} = \frac{S' - \mu_{S'}}{\sigma_{S'}} \cdot \gamma + \beta,
	\end{equation}
	where $\mu_{S'}$ and $\sigma_{S'}$ are instance statistics, and $\gamma, \beta$ are learnable parameters. The normalized prior $\tilde{S}$ is then processed by the geometric adapter $\mathcal{F}_{geo}$, consisting of a $1 \times 1 \times 1$ convolution, Batch Normalization, and a ReLU activation:
	\begin{equation}
		F_{geo} = \mathcal{F}_{geo}(\tilde{S}) = \text{ReLU}(\text{BN}(\text{Conv}_{1\times1\times1}(\tilde{S}))).
	\end{equation}
	This operation maps the geometric scalar field into a high-dimensional semantic space $F_{geo} \in \mathbb{R}^{C_{ad} \times D \times H \times W}$. Finally, the spatial attention map $M_{spatial}$ is derived by aggregating the magnitude of these geometric features:
	\begin{equation}
		M_{spatial}(v) = \text{Norm}\left( \frac{1}{C_{ad}} \sum_{c=1}^{C_{ad}} |F_{geo}^{(c)}(v)| \right),
	\end{equation}
	where $\text{Norm}(\cdot)$ denotes Min-Max normalization to scale the weights to $[0, 1]$. By learning this mapping, the network can dynamically suppress noise in the coarse shape prior and highlight the true anatomical boundaries, effectively acting as a soft, self-calibrated geometric gate.
	
	The Anatomical Weighted Pooling (AWP) operation reformulates the global pooling as a surface integral on the anatomical manifold. It aggregates the input features $X$ using these spatial weights to produce a geometry aware channel descriptor $z' \in \mathbb{R}^{C}$. The $c$-th element of $z'$ is computed as:
	\begin{equation}
		z'_{c} = \frac{\sum_{v \in \Omega} X_{c}(v) \cdot M_{spatial}(v)}{\sum_{v \in \Omega} M_{spatial}(v) + \epsilon},
	\end{equation}
	where $\Omega$ denotes the spatial domain and $\epsilon$ is a stability term. Unlike standard global pooling, $z'$ encodes the channel activations specifically within the anatomical boundary zones. Finally, similar to standard attention mechanisms, $z'$ is processed by a Multi-Layer Perceptron (MLP) bottleneck to generate channel-wise weights $s \in \mathbb{R}^{C}$, which are used to recalibrate the original features:
	\begin{equation}
		s = \mathrm{Sigmoid}\left(W_2 \cdot \mathrm{ReLU}(W_1 \cdot z')\right),
	\end{equation}
	\begin{equation}
		X_{\text{out}}(v) = X(v) \odot s,
	\end{equation}
	where $W_1$ and $W_2$ denote the weights of the dense layers. By integrating SDMAA into the skip connections, we force the network to prioritize features that are spatially consistent with the underlying dental anatomy, thereby effectively suppressing artifacts in ambiguous regions.
	
	\subsection{Loss Function and Deep Supervision}
	\label{subsec:loss}
	
	To facilitate effective model training, particularly given the severe class imbalance inherent in the 39-class dental segmentation task characterized by negligible tooth volume compared to the background, we minimize a composite objective function. This objective, denoted as $\mathcal{L}_{comb}$, integrates the voxel-wise Multi-class Cross-Entropy loss with the region-based Soft Dice loss. The Cross-Entropy loss $\mathcal{L}_{CE}$ is employed to penalize voxel-level classification errors:
	\begin{equation}
		\mathcal{L}_{CE} = - \frac{1}{|\Omega|} \sum_{v \in \Omega} \sum_{c=1}^{C} g_{v,c} \log(p_{v,c}),
	\end{equation}
	where $\Omega$ represents the set of valid voxels, $C$ is the total number of classes, $g_{v,c}$ is the one-hot ground truth label for class $c$ at voxel $v$, and $p_{v,c}$ is the predicted probability.
	
	Simultaneously, to directly optimize the segmentation overlap and mitigate the dominance of background voxels, we employ the Soft Dice loss $\mathcal{L}_{DC}$:
	\begin{equation}
		\mathcal{L}_{DC} = 1 - \frac{1}{C} \sum_{c=1}^{C} \frac{2 \sum_{v \in \Omega} p_{v,c} g_{v,c}}{\sum_{v \in \Omega} p_{v,c} + \sum_{v \in \Omega} g_{v,c} + \epsilon},
	\end{equation}
	where $\epsilon$ is a smoothing term to prevent division by zero.
	
	The final loss for a single output scale is the weighted sum:
	\begin{equation}
		\mathcal{L}_{comb} = \lambda_{CE} \mathcal{L}_{CE} + \lambda_{DC} \mathcal{L}_{DC},
	\end{equation}
	where $\lambda_{CE}$ and $\lambda_{DC}$ serve as balancing coefficients.
	
	Furthermore, to ensure robust gradient propagation to the earlier layers of the encoder and to force the learning of discriminative features at multiple scales, we adopt a deep supervision strategy. We extract predictions $\hat{Y}_k$ from $K$ different resolution levels of the decoder. The total training objective $\mathcal{L}_{total}$ is computed as:
	\begin{equation}
		\mathcal{L}_{total} = \sum_{k=1}^{K} w_k \cdot \mathcal{L}_{comb}(\hat{Y}_k, Y_k),
	\end{equation}
	where $Y_k$ denotes the ground truth downsampled to the resolution of the $k$-th level, and $w_k$ represents the importance weight assigned to that scale.
	
	\section{Experiments}
	
	% Figure 4
	\begin{figure}[!t]
		\centering
		\includegraphics[width=0.9\columnwidth]{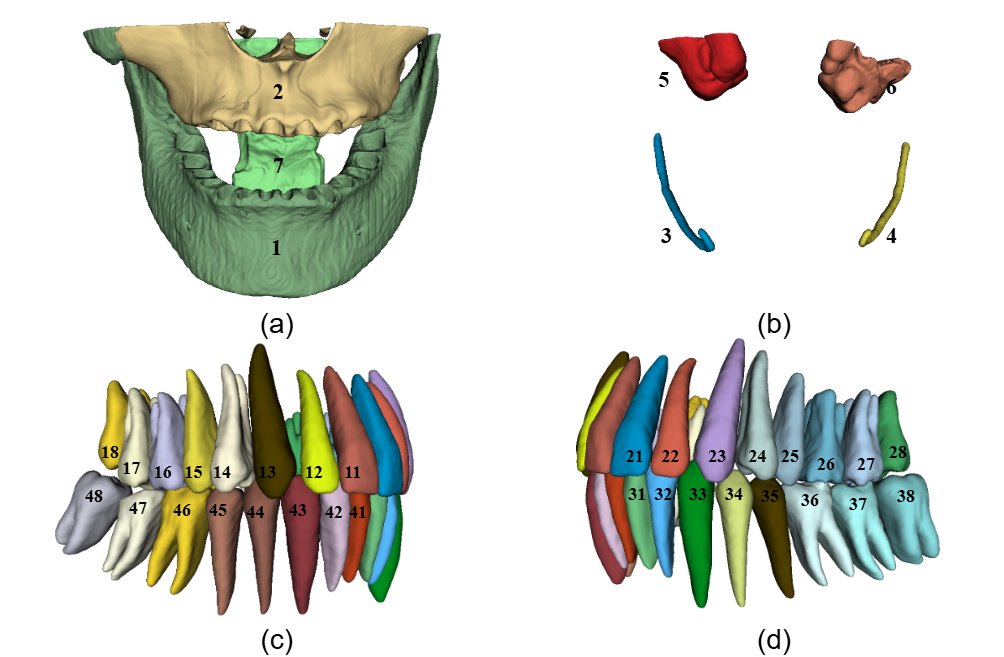}
		\caption{Visualization of the 39 semantic classes defined in our dataset. (a) illustrates  the jawbones and pharynx, while (b) highlights the maxillary sinuses and inferior alveolar canals. (c) and (d) display the 32 individual permanent teeth, indexed according to the standard ISO 3950 notation across the four dental quadrants.}
		\label{fig:dataset}
	\end{figure}
	
	\subsection{Dataset and Preprocessing}
	\label{subsec:dataset}
	
	This study utilizes a large-scale, in-house dental CBCT dataset. The cohort consists of 125 volumetric CBCT scans collected from clinical routine, covering a wide range of anatomical variations and dental conditions. The study protocol was conducted in accordance with the principles of the Declaration of Helsinki, received approval from the Institutional Review Board (No. IIT [2025] LLS No. 482-1), and made use of irreversibly anonymized participant data under a waiver of informed consent granted by the ethics committee. The raw volumes are stored in DICOM format with an isotropic voxel spacing of $0.20$ mm. The spatial resolution varies across subjects, typically centering around $651 \times 651 \times 451$ voxels. We randomly partitioned the dataset at the patient level into a training set of $ 100 $ cases and a testing set of $ 25 $ cases. In addition, we collected CBCT data from the publicly available large-scale CBCT tooth segmentation dataset provided by Cui \textit{et al}.\cite{cui2022fully} as an external test set to evaluate the performance of the proposed model. This large-scale dataset is used for tooth and alveolar bone segmentation to assist in orthodontic diagnosis and dental implant surgery. We randomly selected $20$ cases as the final test set. The original images were in NIFTI format with a voxel spacing of $0.25$ mm and a resolution ranging from $320 \times 320 \times 200$ to $400 \times 400 \times 400$ voxels.
	
	\subsubsection{Annotations}
	To acquire the ground truth labels, the CBCT scans were manually annotated by two dentists with at least $3$ years of clinical experience and validated by a dental expert with more than $5$ years of clinical experience in the orthodontic field. The annotation was performed via Materialise Mimics software (version 21.0). As shown in Fig. \ref{fig:dataset}, the dataset defines 39 distinct semantic classes.
	
	\subsubsection{Preprocessing and Augmentation}
	To standardize the inputs and enhance model generalization, we applied a systematic pipeline. First, intensity values were normalized using Z-score normalization, ensuring a distribution with zero mean and unit variance based on statistics from the training set. Given the high resolution of CBCT volumes, we subsequently employed a patch-based training strategy where volumes were randomly cropped into patches of size $128 \times 224 \times 224$ voxels. Furthermore, to mitigate overfitting and simulate diverse scanning orientations, we implemented data augmentation strategies comprising random rotations within a range of $\pm 15^{\circ}$ and random flipping along the axes.
	
	\subsection{Evaluation Metrics}
	To strictly quantify segmentation performance, we employed four standard metrics divided into two categories: region overlap metrics and boundary precision metrics.
	
	\subsubsection*{1) Region Based Metrics}
	We utilize the DSC and Sensitivity (SEN) to measure the volumetric overlap between the prediction $\mathcal{P}$ and the ground truth $\mathcal{G}$:
	\begin{align}
		\text{DSC} &= \frac{2|\mathcal{P} \cap \mathcal{G}|}{|\mathcal{P}| + |\mathcal{G}|}, \quad 
		\text{SEN} = \frac{|\mathcal{P} \cap \mathcal{G}|}{|\mathcal{G}|},
	\end{align}
	where $|\cdot|$ denotes the number of voxels in the region.
	
	\subsubsection*{2) Boundary Based Metrics}
	To assess the geometric accuracy of the contours, we employ the 95\% Hausdorff Distance (HD95) and the Average Symmetric Surface Distance (ASSD). Let $\partial \mathcal{P}$ and $\partial \mathcal{G}$ denote the surface point sets of the prediction and ground truth, respectively:
	\begin{align}
		\text{HD95} &= \max \left( h_{95}(\partial \mathcal{P}, \partial \mathcal{G}), h_{95}(\partial \mathcal{G}, \partial \mathcal{P}) \right), \\[6pt]
		\text{ASSD} &= \frac{1}{|\partial \mathcal{P}| + |\partial \mathcal{G}|} \Bigg( \sum_{p \in \partial \mathcal{P}} d(p, \partial \mathcal{G}) + \sum_{g \in \partial \mathcal{G}} d(g, \partial \mathcal{P}) \Bigg),
	\end{align}
	where $d(\cdot,\cdot)$ represents the shortest Euclidean distance, and $h_{95}$ indicates the $95^{th}$ percentile of the distance set, which eliminates the impact of small outliers.
	
	\subsection{Implementation Details}
	\label{subsec:implementation}
	
	The proposed AACNet was implemented using PyTorch and trained on a single NVIDIA A100 GPU with 80 GB of memory.
	
	\subsubsection{Training Configuration}
	We employed Stochastic Gradient Descent (SGD) with a momentum of 0.99 and weight decay of $1 \times 10^{-4}$. The initial learning rate was set to 0.01 and decayed via a polynomial schedule $\eta_t = \eta_0 (1 - t/T)^{0.9}$ over 500 epochs. The batch size was set to 2 due to the memory constraints of 3D volumetric processing. The loss balancing coefficients were empirically set to $\lambda_{CE}=1.0$ and $\lambda_{DC}=1.0$. Deep supervision weights were set to $w_k = \{1/2^k\}$ for the auxiliary scales.
	
	\subsubsection{Inference Strategy}
	During inference, we adopted a sliding window strategy with a window size of $128 \times 224 \times 224$ and an overlap of $50\%$ to ensure consistent predictions at patch boundaries. The final segmentation map was reconstructed by stitching the probability maps and applying an argmax operation.
	
	% Figure 5
	\begin{figure*}[!t]
		\centering
		\includegraphics[width=1\textwidth]{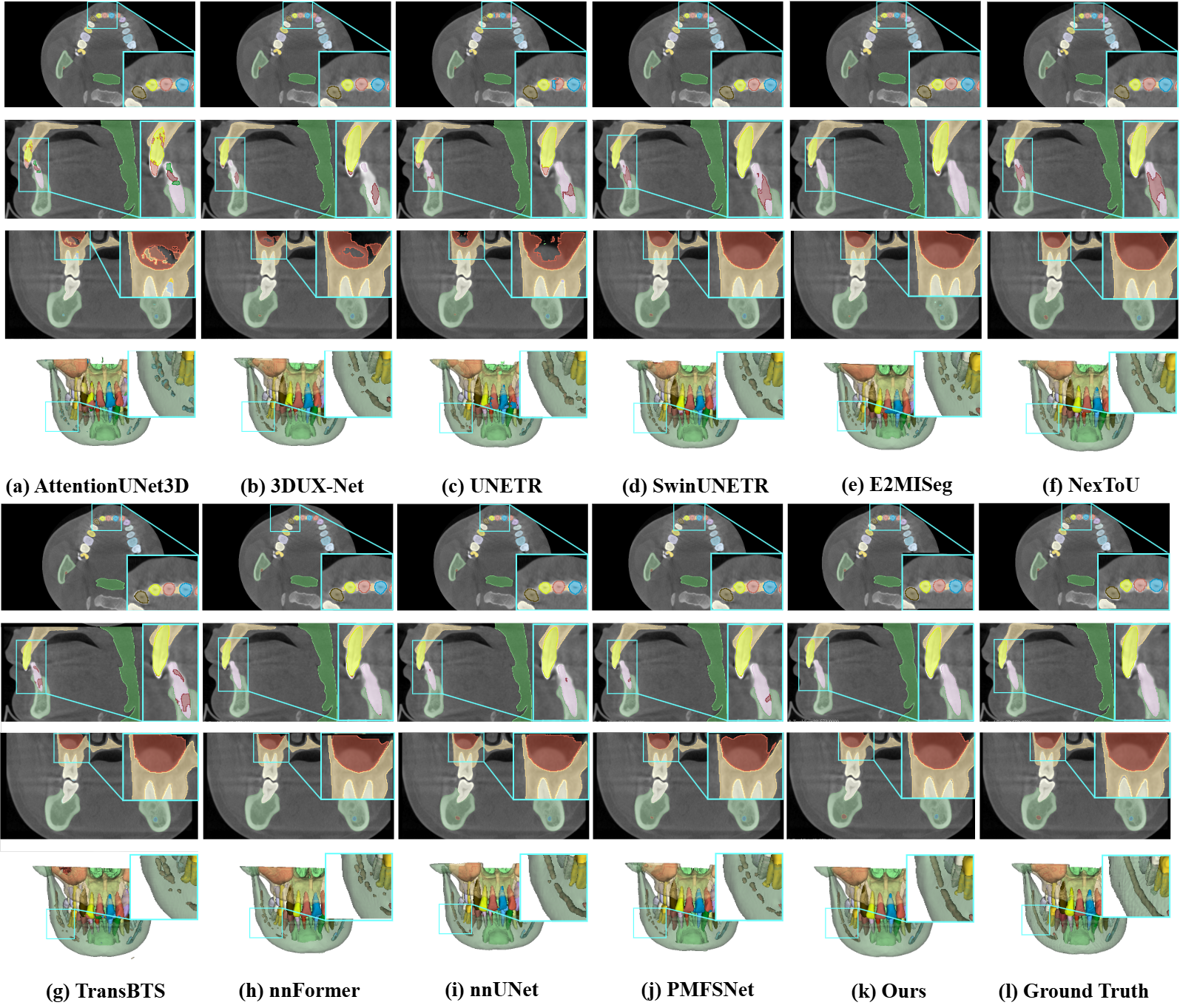}
		\caption{Qualitative comparison of segmentation results on the internal dataset. The proposed AACNet, shown in (k), demonstrates superior fine-grained recovery of root apices and inter-dental boundaries compared to competing state-of-the-art methods. As evidenced by the zoomed regions, our model effectively resolves local ambiguities in high-resolution scans where other baselines fail.} 
		\label{fig:result_1}
	\end{figure*}
	
	\begin{figure*}[!t]
		\centering
		\includegraphics[width=1\textwidth]{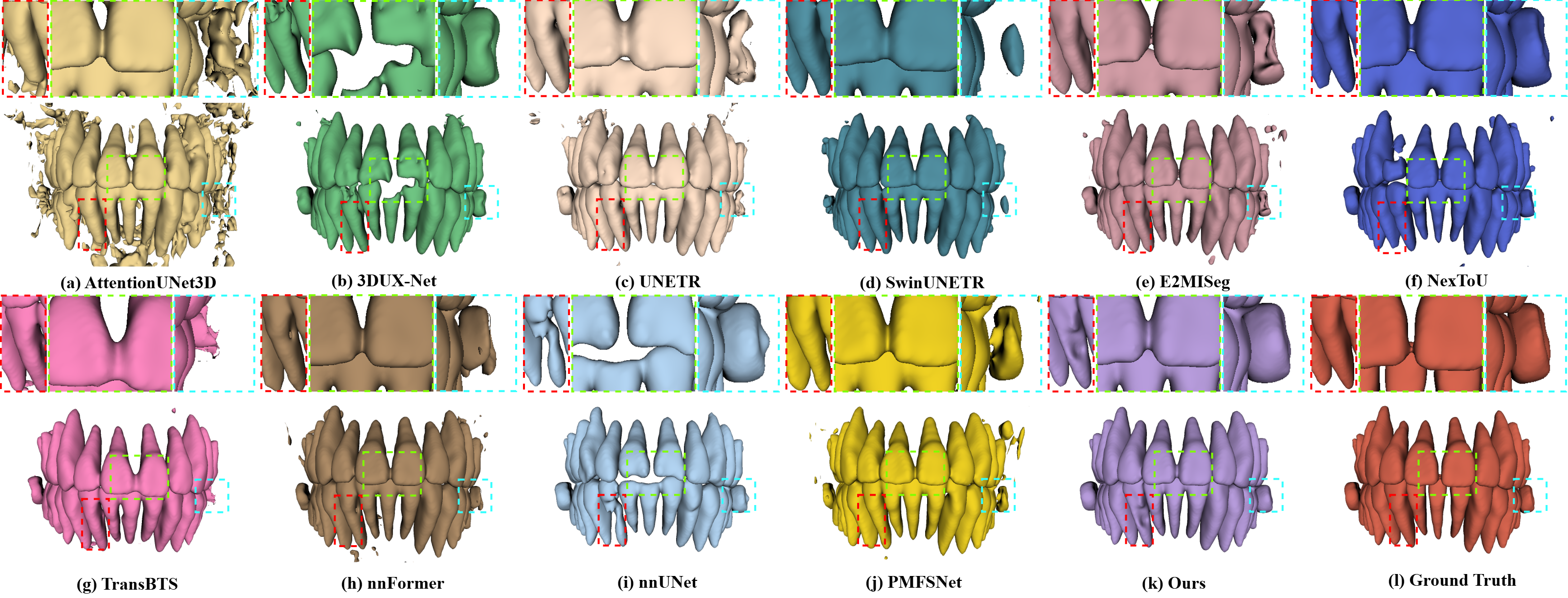}
		\caption{Qualitative assessment of generalization capability on the external dataset. Despite the domain shift caused by different scanner protocols, AACNet maintains smooth, topologically plausible tooth structures. In contrast, baseline methods (e.g., Attention U-Net, SwinUNETR) exhibit jagged contours and island artifacts due to intensity variations, highlighting the robustness of our geometry constrained approach.} 
		\label{fig:result_nc}
	\end{figure*}
	
	\begin{table*}[!t]
		\centering
		\caption{Quantitative Comparison of Segmentation Accuracy and Generalization Assessment on Internal and External Datasets.}
		\label{tab:performance}
		\setlength{\tabcolsep}{2pt} 
		\renewcommand{\arraystretch}{1.35}
		\resizebox{\textwidth}{!}{
			\begin{tabular}{l r@{ $\pm$ }l r@{ $\pm$ }l r@{ $\pm$ }l r@{ $\pm$ }l c r@{ $\pm$ }l r@{ $\pm$ }l r@{ $\pm$ }l r@{ $\pm$ }l}
				\toprule
				\multirow{3}{*}{Method} & \multicolumn{8}{c}{Internal Dataset } & & \multicolumn{8}{c}{External Dataset} \\
				\cmidrule(lr){2-9} \cmidrule(lr){11-18}
				
				& \multicolumn{2}{c}{DSC (\%) $\uparrow$} & 
				\multicolumn{2}{c}{HD95 (mm) $\downarrow$} & 
				\multicolumn{2}{c}{ASSD (mm) $\downarrow$} & 
				\multicolumn{2}{c}{SEN (\%) $\uparrow$} &
				& \multicolumn{2}{c}{DSC (\%) $\uparrow$} & 
				\multicolumn{2}{c}{HD95 (mm) $\downarrow$} & 
				\multicolumn{2}{c}{ASSD (mm) $\downarrow$} & 
				\multicolumn{2}{c}{SEN (\%) $\uparrow$} \\
				\midrule
				
				AttentionUNet3D\cite{oktay2018attention}
				& 83.10 & 2.45 & 96.20 & 21.33 & 18.63 & 5.12 & 83.87 & 2.89 &
				& 50.44 & 10.15 & 184.02 & 45.22 & 60.50 & 16.34 & 80.26 & 7.12 \\
				
				3DUX-Net \cite{lee20223d}
				& 87.34 & 1.82 & 6.69 & 1.45 & 2.20 & 0.56 & 86.70 & 2.01 &
				& 84.51 & 1.94 & 75.13 & 19.56 & 9.93 & 2.87 & 86.41 & 2.15 \\
				
				UNETR \cite{hatamizadeh2022unetr}
				& 85.55 & 2.10 & 14.54 & 4.23 & 3.53 & 1.02 & 85.05 & 2.54 &
				& 80.05 & 3.25 & 122.89 & 36.45 & 17.77 & 4.88 & 84.46 & 3.67 \\
				
				SwinUNETR \cite{cao2022swin}
				& 87.76 & 1.65 & 6.49 & 1.32 & 2.02 & 0.48 & 87.17 & 1.88 &
				& 84.45 & 1.76 & 45.91 & 12.44 & 7.47 & 1.62 & 84.89 & 1.95 \\
				
				TransBTS \cite{wang2021transbts}
				& 84.29 & 1.89 & 33.84 & 1.71 & 7.31 & 1.32 & 84.20 & 2.11 &
				& 83.26 & 2.05 & 72.96 & 17.82 & 10.06 & 2.45 & 84.14 & 2.24 \\
				
				E2MISeg \cite{jiang2026e2miseg}  
				& 86.78 & 1.72 & 10.06 & 1.16 & 2.60 & 0.73 & 87.52 & 1.68 &
				& 84.13 & 1.83 & 59.86 & 14.28 & 9.22 & 2.21 & 85.85 & 1.79 \\
				
				NexToU  \cite{shi2023nextou}
				& 87.99 & 1.32 & 9.52 & 1.03 & 2.58 & 0.65 & 88.17 & 1.58 &
				& 85.49 & 1.45 & 41.42 & 9.67 & 7.53 & 1.74 & 85.15 & 1.62 \\
				
				nnFormer  \cite{zhou2023nnformer}
				& 85.82 & 1.95 & 9.06 & 2.15 & 2.90 & 0.75 & 86.57 & 2.23 &
				& 82.85 & 2.88 & 99.39 & 23.56 & 13.12 & 3.92 & 86.37 & 2.45 \\
				
				nnUNet \cite{isensee2021nnu}
				& 88.44 & 1.35 & 5.19 & 0.98 & 1.76 & 0.34 & 87.93 & 1.56 &
				& 85.77 & 1.28 & 3.22 & 1.85 & \textbf{1.13} & \textbf{0.42} & 84.46 & 1.45 \\
				
				PMFSNet \cite{zhong2025pmfsnet}
				& 54.37 & 12.50 & 126.32 & 35.41 & 50.75 & 10.20 & 47.52 & 14.10 &
				& 79.85 & 11.85 & 116.08 & 36.12 & 14.91 & 9.85 & 80.78 & 13.50 \\
				
				\midrule
				\textbf{AACNet (Ours)} 
				& \textbf{90.17} & \textbf{1.12} 
				& \textbf{3.63} & \textbf{0.76} 
				& \textbf{1.54} & \textbf{0.28} 
				& \textbf{90.00} & \textbf{1.25} &
				& \textbf{87.19} & \textbf{0.96} 
				& \textbf{2.19} & \textbf{0.82} 
				& 1.23 & 0.25 
				& \textbf{88.56} & \textbf{1.18} \\
				\bottomrule
				\multicolumn{18}{l}{\scriptsize Results are reported as Mean $\pm$ Standard Deviation. The best results are highlighted in \textbf{bold}.}
			\end{tabular}
		}
	\end{table*}
	
	\subsection{Quantitative Evaluations}
	\label{subsec:quantitative}
	To provide a rigorous assessment, we benchmarked AACNet against eight state-of-the-art methods, encompassing generic CNN baselines including Attention U-Net, nnUNet, and nnFormer; Transformer-based architectures such as UNETR, SwinUNETR, 3DUX-Net, and TransBTS; and recent task-specific networks designed for boundary and topological modeling, namely E2MISeg and NexToU.
	Additionally, we included the lightweight PMFSNet for CBCT segmentation. The comparative results on both the internal dataset and the external dataset are summarized in Table \ref{tab:performance}.
	
	\subsubsection{Performance on Internal Dataset}
	In terms of volumetric overlap metrics on the internal cohort, AACNet demonstrates a significant performance improvement. As evidenced in Table~\ref{tab:performance}, our method achieved a DSC of 90.17\%, distinguishing itself as the only model to surpass the 90\% threshold. This represents a substantial margin over strong baselines like nnUNet at 88.44\% and the topology-aware NexToU at 87.99\%. Crucially, regarding SEN, AACNet reached 90.00\%, outperforming the edge-focused E2MISeg by nearly 2.5\%. This confirms that the AGBR effectively recovers subtle root signals that are often suppressed by standard self-attention mechanisms.
	
	\subsubsection{Generalization Capability on External Dataset}
	To evaluate the robustness of our framework against domain shifts caused by different scanner protocols and resolutions, we conducted an external validation on external dataset. As shown in the right section of Table~\ref{tab:performance}, the performance gap between AACNet and competing methods widens significantly in this challenging scenario. Regarding robustness to domain shift, while generic models experienced severe performance degradation, exemplified by Attention U-Net where the HD95 spiked to 184.02~mm due to complete localization failures, AACNet maintained high stability with a DSC of 88.56\% and a SEN of 88.56\%. This corresponds to a 4.1\% improvement in DSC over the second-best method, nnUNet, which achieved 84.46\%. Most notably, in terms of geometric consistency, AACNet achieved an HD95 of 2.19~mm and an ASSD of 1.23~mm on the external set. This result is critical, it suggests that while intensity distributions vary across scanners affecting texture-based CNNs, the underlying anatomical topology remains invariant. By embedding implicit geometric manifold constraints via the SDMAA module, our network learns to rely on stable shape priors rather than fluctuating texture cues, thereby ensuring superior generalization on unseen data.
	
	\subsubsection{Boundary Precision Analysis}
	The superiority of AACNet is consistent across both datasets.
	In internal dataset, we achieved an HD95 of 3.63~mm, yielding a substantial error reduction of over 60\% compared to task-specific competitors like E2MISeg with 10.06~mm and NexToU with 9.52~mm. This confirms that explicit geometric priors, rather than just feature-level edge enhancement or graph-based constraints, are essential for resolving adhesion artifacts in dental CBCT.
	
	\subsection{Qualitative Results Analysis}
	
	\subsubsection{Internal Validation}
	As illustrated in Fig. \ref{fig:result_1}, the visual comparisons are consistent with the quantitative results.
	A detailed inspection of the zoomed-in regions reveals distinct failure modes among the competing methods.
	Generic models such as Attention U-Net and UNETR exhibit severe under-segmentation, frequently failing to capture the fine details of root apices.
	More importantly, despite their specialized designs, both E2MISeg and NexToU fail to fully address the core challenge of occlusal adhesion.
	E2MISeg produces noisy, open boundaries between teeth, while NexToU exhibits localized fusion between the upper and lower arches.
	Even the robust nnUNet generates segmentation masks with visible jagged edges.
	Conversely, the AACNet generates segmentation masks with high anatomical fidelity.
	Benefiting from the AGBR, our model yields sharp, distinct boundaries that effectively eliminate adhesion artifacts.
	Furthermore, the SDMAA module ensures that the segmented surfaces are smooth and consistent with biological priors, surpassing the geometric precision of all competitors.
	
	\subsubsection{External Generalization Analysis}
	As illustrated in Fig.~\ref{fig:result_nc}, we visualize the segmentation results on the external test set, characterized by distinct noise patterns and contrast levels relative to the training data.
	\begin{itemize}
		\item \noindent Handling Unseen Variations: As observed in the zoomed-in regions, baseline methods such as TransBTS and SwinUNETR exhibit jagged contours and island artifacts where the network fails to distinguish teeth from the alveolar bone due to domain differences.
		\item \noindent Topology Preservation: In contrast, AACNet generates smooth, biologically plausible tooth roots that closely match the ground truth.
		The AGBR module successfully filters out the epistemic uncertainty introduced by the unfamiliar image statistics, while the SDMAA ensures that even in low-contrast regions, the predicted shapes adhere to the learned dental manifold.
		This qualitative evidence strongly supports the quantitative findings that our coarse-to-fine, anatomy-aware strategy offers superior clinical viability for multi-center deployment.
	\end{itemize}
	
	In contrast, by integrating the SDM-based shape prior via our SDMAA module, AACNet achieved an HD95 of 3.63~mm, a significant error reduction of over 60\% compared to these task-specific competitors.
	This confirms that explicit geometric priors, rather than just feature-level edge enhancement or graph-based constraints, are essential for resolving adhesion artifacts in dental CBCT.
	
	\subsection{Clinical Application Analysis}
	\label{subsec:clinical_analysis}
	
	To bridge the gap between algorithmic development and clinical practice, we validated AACNet on two critical preoperative assessment tasks: maxillary sinus proximity analysis and risk assessment for the inferior alveolar canal (IAC).
	These tasks require not only segmentation accuracy but also high geometric fidelity to prevent intraoperative complications.
	We quantify clinical reliability by comparing automated measurements against expert manual benchmarks.
	Let $D_{auto}$ denote the shortest Euclidean distance computed from the AACNet-generated 3D meshes, and $D_{ref}$ represent the ground truth distance measured manually by a senior radiologist on raw CBCT slices.
	The measurement discrepancy is defined as $\Delta E = |D_{auto} - D_{ref}|$.
	
	\subsubsection{Sinus Proximity Risk Assessment}
	In implant dentistry and maxillofacial surgery, procedures such as posterior maxillary implant placement and maxillary sinus lifts are common.
	A precise preoperative assessment of the spatial relationship between tooth roots and the sinus floor is essential for evaluating implant feasibility and, critically, for avoiding severe complications like sinus membrane perforation.
	
	Analysis Strategy: This analysis focuses exclusively on the posterior maxillary teeth, specifically premolars and molars.
	This selection is based on clear clinical considerations: anatomically, the root apices of these teeth have the most intimate relationship with the maxillary sinus floor, representing the highest risk regions for sinus-related surgeries.
	
	Results and Discussion: The quantitative results are presented in Table \ref{tab:Sinus}.
	To ensure clinical precision, we computed the shortest Euclidean distance directly from the reconstructed 3D surface meshes of the tooth roots and the maxillary sinus floor.
	Unlike discrete voxel counting, this mesh-based approach leverages the continuous geometric manifold recovered by AACNet, enabling accurate sub-millimeter measurements without the need for manual calibration.
	Notably, as detailed in Table \ref{tab:Sinus}, the system successfully identified clinically confirmed root protrusions, achieving $\Delta E$ of 0.21~mm.
	This high concordance confirms that the AACNet-derived measurements are reliable for identifying high-risk protrusion cases in preoperative planning.
	
	\begin{table}[!t]
		\centering
		\caption{Transposed Quantitative Analysis of Maxillary Sinus Proximity.}
		\label{tab:Sinus}
		\renewcommand{\arraystretch}{1.15}
		\setlength{\tabcolsep}{1.5pt}
		
		\resizebox{\columnwidth}{!}{%
			\begin{tabular}{l ccccc ccccc}
				\toprule
				Metric & 14 & 15 & 16 & 17 & 18 & 24 & 25 & 26 & 27 & 28 \\
				\midrule
				$D_{auto}$ & 3.13 & 0.08 & 0.00 & 0.00 & 0.15 & 0.29 & 0.00 & 0.08 & 0.20 & 1.13 \\
				$D_{ref}$  & 3.26 & 0.14 & -0.46 & -0.62 & 0.22 & 0.48 & -0.49 & 0.09 & 0.22 & 1.18 \\
				$\Delta E$ & 0.13 & 0.06 & 0.46\rlap{$^{*}$} & 0.62\rlap{$^{*}$} & 0.07 & 0.19 & 0.49\rlap{$^{*}$} & 0.01 & 0.02 & 0.06 \\
				\midrule
				Average Error & \multicolumn{10}{r}{\textbf{0.21 mm}} \\
				\bottomrule
				\multicolumn{11}{l}{\scriptsize $^{*}$ indicates successful identification of sinus protrusion risk.}
			\end{tabular}%
		}
	\end{table}
	
	% Table 3
	\begin{table}[!t]
		\centering
		\caption{Transposed Quantitative Analysis of IAC Proximity.}
		\label{tab:Inferior}
		
		\renewcommand{\arraystretch}{1.15}
		\setlength{\tabcolsep}{1.5pt}
		
		\resizebox{\columnwidth}{!}{%
			\begin{tabular}{l ccccc ccccc}
				\toprule
				% 去除了表头的加粗
				Metric & 34 & 35 & 36 & 37 & 38 & 44 & 45 & 46 & 47 & 48 \\
				\midrule
				% 去除了第一列和内容的加粗
				$D_{auto}$ & 5.60 & 5.20 & 6.56 & 2.70 & 1.30 & 4.63 & 4.30 & 5.88 & 3.00 & 1.77 \\
				$D_{ref}$  & 6.02 & 5.78 & 6.68 & 3.01 & 1.45 & 4.81 & 4.72 & 6.16 & 3.09 & 2.01 \\
				$\Delta E$ & 0.42 & 0.58 & 0.12 & 0.31 & 0.15 & 0.18 & 0.42 & 0.28 & 0.09 & 0.24 \\
				\midrule
				% 标识放入第一列，数值放入多列合并，并且只加粗数值
				Average Error & \multicolumn{10}{r}{\textbf{0.28 mm}} \\
				\bottomrule
			\end{tabular}%
		}
	\end{table}
	
%	% Figure 6
%	\begin{figure}[!t] 
%		\centering
%		\includegraphics[width=0.55\linewidth]{sinus_6.png}
%		\caption{Visualization of the Maxillary Sinus Proximity Analysis. The segmented maxillary sinuses (semi-transparent red) and posterior tooth roots (multi-colored) are reconstructed as high-fidelity meshes. Colored lines quantify the shortest Euclidean distance (\(D_{auto}\)) from each root apex to the sinus floor, providing an intuitive risk map for potential sinus membrane perforation during implant surgery.}
%		\label{fig:sinus_analysis}
%	\end{figure}
	
	\subsubsection{Inferior Alveolar Canal Proximity Analysis}
	The IAC is a critical anatomical structure within the mandible.
	Accurate assessment of the root-to-IAC spatial relationship is imperative during lower jaw surgeries, particularly third molar extraction and implant placement, to prevent iatrogenic nerve injury.
	
	Analysis Strategy: This analysis analogously focuses on the posterior mandibular teeth, including premolars and molars, as they are anatomically closest to the IAC and pose the highest risk of nerve injury during surgery.
	
	Results and Discussion: The quantitative metrics for this analysis are presented in Table \ref{tab:Inferior}.
	Consistent with our sinus proximity analysis, we utilized the reconstructed high-fidelity meshes for automated measurement to avoid voxel discretization errors.
	$D_{auto}$ was computed using efficient cKDTree-based nearest point queries to ensure exact Euclidean distance calculation without manual calibration.
	The results demonstrate a high degree of concordance with the manual reference measurements $D_{ref}$, yielding $\Delta E$ of only 0.28~mm across all posterior teeth.
	This sub-millimeter accuracy confirms that our model successfully preserves the critical safety margins required for third molar extractions.
	
	Conclusion: The two preceding analyses demonstrate that the AACNet framework not only achieves high precision segmentation but also produces geometric measurements that are highly consistent with manual clinical assessment.
	Our method provides clinicians with an intuitive and quantitatively precise tool for risk evaluation, showing strong potential for clinical translation in dental surgical planning.
	
	\subsection{Ablation Study}
	\label{subsec:ablation}
	
	To verify the contribution of each component within the proposed AACNet framework, we conducted comprehensive ablation studies on the test set.
	The quantitative results regarding module ablation and attention comparison are summarized in Table \ref{tab:ablation} and Table \ref{tab:ablation_sdmaa}, respectively.
	
	\subsubsection{Effectiveness of Key Modules}
	As shown in Table \ref{tab:ablation}, the baseline model, defined as a standard residual U-Net without the proposed modules, achieved a DSC of 84.93\% and an HD95 of 6.56~mm.
	This relatively high error rate is primarily attributed to the model's inability to distinguish unclear boundaries in occluded regions.
	
	Impact of AGBR: By integrating the AGBR into the baseline, the DSC increased by 1.42\% to 86.35\%, and the ASSD decreased from 2.32~mm to 2.04~mm.
	This improvement indicates that explicitly modeling boundary uncertainty allows the network to focus its learning capacity on the most challenging inter-arch regions, thereby reducing confusion between adjacent teeth.
	
	Impact of SDMAA: The introduction of the SDMAA module yielded a substantial performance boost.
	Compared to the baseline, the configuration incorporating SDMAA improved the DSC by 4.20\% to reach 89.13\% and significantly reduced the HD95 by 35.3\% to 4.24~mm.
	This significant reduction in boundary error confirms that incorporating explicit shape priors into the feature learning process effectively constrains the segmentation output to a plausible anatomical topology, preventing the generation of jagged or irregular boundaries.
	
	Synergistic Effect: The complete AACNet framework, which combines both AGBR and SDMAA, achieved the best performance across all metrics.
	The DSC reached 90.17\% and the SEN peaked at 90.00\%, demonstrating that the two modules are complementary: AGBR resolves local ambiguity at the voxel level, while SDMAA ensures global structural consistency.
	\subsubsection{Comparison of Attention Mechanisms}
	To further validate the superiority of our proposed anatomical attention mechanism, we compared SDMAA with two widely used general-purpose attention modules: the Squeeze and Excitation block, abbreviated as SE \cite{hu2018squeeze}, and the CBAM.
	As detailed in Table \ref{tab:ablation_sdmaa}, while adding SE or CBAM to the baseline improved the DSC to 88.86\% and 88.04\% respectively, our SDMAA module outperformed both.
	Crucially, in terms of boundary precision, SDMAA achieved an HD95 of 4.24~mm, which is significantly lower than the 5.66~mm and 4.90~mm recorded by SE and CBAM, respectively.
	This performance gap highlights the limitation of standard GAP used in SE and CBAM, which discards spatial information.
	In contrast, the Anatomical Weighted Pooling of SDMAA effectively preserves spatial cues related to anatomical surfaces, enabling the network to learn more discriminative features for boundary delineation.

	\begin{table}[!t]
		\centering
		\caption{Ablation study on the effectiveness of key modules.}
		\label{tab:ablation}
		\renewcommand{\arraystretch}{1.35}
		\setlength{\tabcolsep}{2pt} 
		\resizebox{\columnwidth}{!}{%
			\begin{tabular}{l r@{ $\pm$ }l r@{ $\pm$ }l r@{ $\pm$ }l r@{ $\pm$ }l}
				\toprule
				Method & \multicolumn{2}{c}{DSC (\%) $\uparrow$} & 
				\multicolumn{2}{c}{HD95 (mm) $\downarrow$} & 
				\multicolumn{2}{c}{ASSD (mm) $\downarrow$} & 
				\multicolumn{2}{c}{SEN (\%) $\uparrow$} \\
				\midrule
				Baseline          
				& 84.93 & 2.15 & 6.56 & 1.85 & 2.32 & 0.65 & 81.69 & 2.40 \\
				
				Baseline + AGBR   
				& 86.35 & 1.92 & 6.24 & 1.60 & 2.04 & 0.58 & 83.09 & 2.10 \\
				
				Baseline + SDMAA  
				& 89.13 & 1.45 & 4.24 & 0.95 & 1.69 & 0.42 & 88.90 & 1.55 \\
				
				\textbf{AACNet (Ours)} 
				& \textbf{90.17} & \textbf{1.12} 
				& \textbf{3.63} & \textbf{0.76} 
				& \textbf{1.54} & \textbf{0.28} 
				& \textbf{90.00} & \textbf{1.25} \\
				\bottomrule
				\multicolumn{9}{l}{\scriptsize Assessing the incremental contributions of AGBR and SDMAA to the baseline.}
			\end{tabular}%
		}
	\end{table}
	
	\begin{table}[!t]
		\centering
		\caption{Comparison of attention mechanisms.}
		\label{tab:ablation_sdmaa}
		\renewcommand{\arraystretch}{1.35}
		\setlength{\tabcolsep}{2pt}
		\resizebox{\columnwidth}{!}{%
			\begin{tabular}{l r@{ $\pm$ }l r@{ $\pm$ }l r@{ $\pm$ }l r@{ $\pm$ }l}
				\toprule
				Method & \multicolumn{2}{c}{DSC (\%) $\uparrow$} & \multicolumn{2}{c}{HD95 (mm) $\downarrow$} & \multicolumn{2}{c}{ASSD (mm) $\downarrow$} & \multicolumn{2}{c}{SEN (\%) $\uparrow$} \\
				\midrule
				Baseline + SE
				& 88.86 & 1.58 & 5.66 & 1.20 & 1.75 & 0.45 & 88.85 & 1.60 \\
				
				Baseline + CBAM
				& 88.04 & 1.65 & 4.90 & 1.10 & 1.95 & 0.50 & 87.37 & 1.72 \\
				
				\textbf{Baseline + SDMAA}
				& \textbf{89.13} & \textbf{1.45} & \textbf{4.24} & \textbf{0.95} & \textbf{1.69} & \textbf{0.42} & \textbf{88.90} & \textbf{1.55} \\
				\bottomrule
				\multicolumn{9}{l}{\scriptsize Evaluating the proposed SDMAA against SE and CBAM.}
			\end{tabular}%
		}
	\end{table}
	
	\begin{table}[!t]
		\centering
		\caption{Sensitivity Analysis of the Threshold $\tau$ in AGBR.}
		\label{tab:ablation_tau}
		\renewcommand{\arraystretch}{1.35}
		\setlength{\tabcolsep}{3pt}
		\resizebox{\columnwidth}{!}{%
			\begin{tabular}{c c r@{ $\pm$ }l r@{ $\pm$ }l r@{ $\pm$ }l}
				\toprule
				Strategy & $\tau$ Value & \multicolumn{2}{c}{DSC (\%) $\uparrow$} & \multicolumn{2}{c}{HD95 (mm) $\downarrow$} & \multicolumn{2}{c}{ASSD (mm) $\downarrow$} \\
				\midrule
				Global Refinement & 0.00 & 89.25 & 1.35 & 4.15 & 0.92 & 1.78 & 0.35 \\
				\midrule
				\multirow{4}{*}{Gated Refinement}
				& 0.50 & 89.42 & 1.30 & 4.02 & 0.88 & 1.72 & 0.33 \\
				& 0.80 & 89.88 & 1.25 & 3.85 & 0.84 & 1.63 & 0.30 \\
				& 0.90 & 90.05 & 1.18 & 3.72 & 0.79 & 1.58 & 0.29 \\
				& \textbf{0.95} & \textbf{90.17} & \textbf{1.12} & \textbf{3.63} & \textbf{0.76} & \textbf{1.54} & \textbf{0.28} \\
				& 0.99 & 89.31 & 1.41 & 4.10 & 0.95 & 1.75 & 0.38 \\
				\bottomrule
				\multicolumn{8}{l}{\scriptsize We evaluate the impact of the gating threshold on segmentation accuracy.} \\
				\multicolumn{8}{l}{\scriptsize Optimal performance is achieved at $\tau=0.95$.}
			\end{tabular}%
		}
	\end{table}	
	
	\begin{table}[!t]
		\centering
		\caption{Ablation Study on the Design Choices of SDMAA.}
		\label{tab:ablation_sdmaa_internal}
		\renewcommand{\arraystretch}{1.35}
		\setlength{\tabcolsep}{2pt}
		\resizebox{\columnwidth}{!}{%
			\begin{tabular}{l l r@{ $\pm$ }l r@{ $\pm$ }l r@{ $\pm$ }l}
				\toprule
				Guide Source & Pooling & \multicolumn{2}{c}{DSC (\%) $\uparrow$} & \multicolumn{2}{c}{HD95 (mm) $\downarrow$} & \multicolumn{2}{c}{ASSD (mm) $\downarrow$} \\
				\midrule
				Binary Mask ($M$) & AWP & 88.52 & 1.55 & 5.12 & 1.15 & 1.88 & 0.48 \\
				SDM ($\phi$) & GAP & 89.45 & 1.38 & 4.18 & 0.85 & 1.65 & 0.39 \\
				\textbf{SDM} ($\phi$) & \textbf{AWP} & \textbf{90.17} & \textbf{1.12} & \textbf{3.63} & \textbf{0.76} & \textbf{1.54} & \textbf{0.28} \\
				\bottomrule
				\multicolumn{8}{l}{\scriptsize Comparison of geometric guide sources (Mask vs. SDM) and pooling} \\
				\multicolumn{8}{l}{\scriptsize strategies (GAP vs. AWP).}
			\end{tabular}%
		}
	\end{table}
	
	\begin{figure}[!t]
		\centering
		\includegraphics[width=0.95\linewidth]{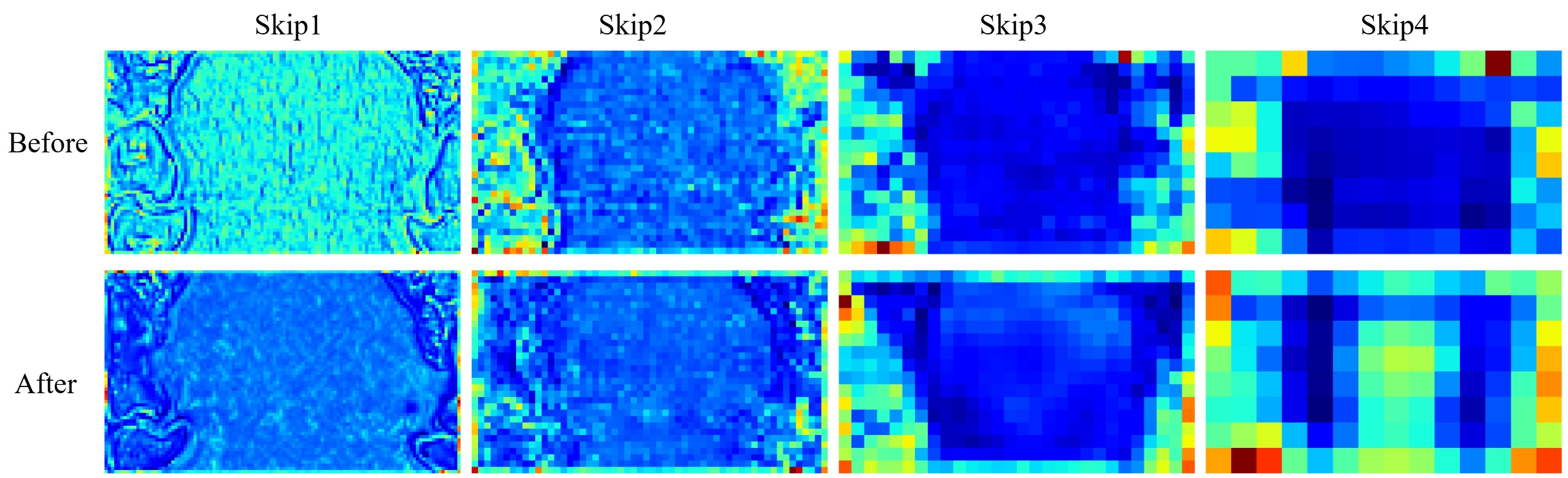} 
		\caption{Visualization of feature activation maps from the decoder's skip connections, labeled Skip 1-4.
			Top Row: The raw features before refinement exhibit diffuse high-frequency noise in non-anatomical regions, visible as cyan/green clutter.
			Bottom Row: The application of SDMAA significantly suppresses these background signals, shifting to deep blue, while maintaining high activations along the dental boundaries.
			This visual contrast illustrates the module's capability to filter out task-irrelevant features based on geometric constraints.}
		\label{fig:feature_vis}
	\end{figure}

	\subsubsection{Sensitivity Analysis of the AGBR Threshold $\tau$}
	We further investigate the impact of the ambiguity threshold $\tau$ on the segmentation performance, as shown in Table \ref{tab:ablation_tau}.
	Setting $\tau=0$ corresponds to a global refinement strategy where all voxels are rectified.
	While this yields a decent DSC of 89.25\%, it introduces computational redundancy and potential noise in high-confidence regions.
	As $\tau$ increases, the module selectively focuses on uncertain boundaries, improving performance.
	The metrics peak at $\tau=0.95$, yielding a DSC of 90.17\% and an HD95 of 3.63~mm, confirming our hypothesis that targeting the hardest voxels is most effective.
	Conversely, an overly strict threshold of $\tau=0.99$ restricts the refinement scope too much, causing performance to drop back to 89.31\%.
	
	\subsubsection{Effectiveness of Geometric Priors and Pooling Strategies}
	To validate the internal design of SDMAA, we conducted an ablation study on the guide map source and pooling mechanism, as summarized in Table \ref{tab:ablation_sdmaa_internal}.
	\begin{itemize} 
		\item \textit{Guide Source:} Using a binary mask yields inferior boundary precision compared to the SDM.
		This demonstrates that the Lipschitz-continuous gradient flow provided by the SDM is crucial for guiding the network towards the anatomical surface.
		\item \textit{Pooling Strategy:} Replacing our AWP with standard GAP results in a performance drop, where the DSC decreases from 90.17\% to 89.45\%.
		This indicates that GAP discards critical spatial information, whereas AWP effectively performs a manifold-constrained integration, preserving the geometric details essential for separating adjacent teeth.
	\end{itemize}
	\subsubsection{Qualitative Analysis of Geometric Refinement}
	To validate the mechanism of the SDMAA module, we visualized the intermediate feature tensors in Fig. \ref{fig:feature_vis}.
	As hypothesized, the standard encoder shown in the top row struggles to differentiate between the dental anatomy and the surrounding alveolar bone, resulting in ambiguous feature representations in the background.
	In contrast, the SDMAA refined features shown in the bottom row demonstrate a strict adherence to the anatomical manifold.
	By integrating the SDM, the module effectively acts as a geometric gate, which penalizes activations that violate the underlying topology.
	This proves that our method does not merely memorize texture, but actively utilizes the implicit shape prior to prune topological noise, thereby sharpening the decision boundary for the subsequent segmentation head.
	\section{Conclusion}
	\label{sec:conclusion}
	In this study, we addressed the persistent challenge of adhesion artifacts in dental CBCT segmentation by proposing the AACNet.
	By synergistically integrating uncertainty modeling with geometric priors, our framework bridges the gap between data-driven feature learning and explicit anatomical constraints.
	Extensive experiments demonstrate that AACNet achieves a new state-of-the-art performance on the internal cohort, with a DSC of 90.17\% and an HD95 of 3.63~mm, while maintaining robust generalization on unseen external data.
	
	We attribute this robustness primarily to the distinct mechanisms of our proposed modules.
	The SDMAA module effectively decouples segmentation from scanner-specific intensity variations by embedding implicit signed distance constraints.
	Since the geometric manifold of dental anatomy remains invariant across devices while texture does not, this geometric prior enables the model to resist domain shifts.
	Complementing this, the AGBR module functions as an entropy-driven filter, specifically targeting high ambiguity contact zones.
	By dynamically rectifying features where the model exhibits high epistemic uncertainty, AGBR ensures that subtle root apices and inter-arch boundaries are preserved, as evidenced by our superior sensitivity metrics.
	
	Despite these advantages, the current two-stage cascaded architecture inherently increases computational demand and inference latency compared to single-stage networks.
	Furthermore, the final precision relies on the coarse localization from the first stage.
	Future work will focus on distilling the knowledge from this cascaded model into a lightweight, single-stage network for real-time deployment and extending the shape prior mechanism to encompass diverse oral pathologies such as cysts and tumors.
	In summary, by ensuring topological consistency and sub-millimeter precision in clinical risk assessments, AACNet establishes a reliable, geometry-aware foundation for the next generation of automated digital dental workflows.
	
	\section{REFERENCES}
	\bibliographystyle{IEEEtran}
	\bibliography{ref}
	
\end{document}